\documentclass[preprint,12pt, a4paper]{elsarticle}

\usepackage{amssymb}
\usepackage{hyperref}
\usepackage{listings}
\usepackage{xcolor}
\usepackage{tabularx}
\usepackage[numbers]{natbib}
\usepackage[utf8]{inputenc}
\usepackage[T1]{fontenc}
\usepackage{amsmath,amssymb,amsfonts}
\usepackage{algorithmic}
\usepackage{graphicx}
\usepackage{textcomp}
\usepackage{hyperref}
\usepackage{xcolor}
\usepackage{multicol}
\usepackage{subcaption}

\def\BibTeX{{\rm B\kern-.05em{\sc i\kern-.025em b}\kern-.08em
    T\kern-.1667em\lower.7ex\hbox{E}\kern-.125emX}}

\def\VPprojectPath{vp_project}

\newcommand{\Fig}[1]{Fig.~\ref{#1}}

\def\StTask{<<Task>>}
\def\StTasks{<<Task>>s}
\def\StEnvironment{<<Environment>>}

\def\StDecAgent{<<DecAgent>>}
\def\StDecAgents{<<DecAgent>>s}
\def\StScenario{<<Scenario>>}
\def\StScenarios{<<Scenario>>s}
\def\StRobot{<<Robot>>}

\def\StTabsaSystem{<<TaBSASystem>>}

\def\StScenarioPlugin{<<ScenarioPlugin>>}
\def\StScenarioPlugins{<<ScenarioPlugin>>s}
\def\StAgentPlugin{<<AgentPlugin>>}
\def\StAgentPlugins{<<AgentPlugin>>s}
\def\StEvalFunction{<<EvalFunction>>}
\def\StEvalFunctions{<<EvalFunction>>s}

\lstdefinestyle{pythonstyle}{
    language=Python,
    basicstyle=\ttfamily\footnotesize, 
    backgroundcolor=\color{gray!10}, 
    frame=single, 
    rulecolor=\color{gray!50}, 
    keywordstyle=\color{blue}, 
    commentstyle=\color{teal}, 
    stringstyle=\color{red}, 
    showstringspaces=false, 
    breaklines=true, 
    morecomment=[l][\color{magenta}]{\#} 
}

\journal{SoftwareX}

\begin{document}
\renewcommand{\labelenumii}{\arabic{enumi}.\arabic{enumii}}

\begin{frontmatter}

\title{TaBSA -- A framework for training and benchmarking algorithms scheduling tasks for mobile robots working in dynamic environments}

\author[wut]{Wojciech Dudek}
\author[wut]{Daniel Giełdowski}
\author[put]{Kamil Młodzikowski}
\author[put]{Dominik Belter}
\author[wut]{Tomasz Winiarski}
\address[wut]{Warsaw University of Technology, Institute of Control and Computation Engineering, Poland, Nowowiejska 15/19, 00-665 Warsaw, Poland, name.surname@pw.edu.pl}
\address[put]{Institute of Robotics and Machine Intelligence, Pozna\'n University of Technology, Pl. Marii Sklodowskiej-Curie 5, PL 60-965 Pozna\'n, Poland, name.surname@put.poznan.pl}

\begin{abstract}
This article introduces a software framework for benchmarking robot task scheduling algorithms in dynamic and uncertain service environments. The system provides standardized interfaces, configurable scenarios with movable objects, human agents, tools for automated test generation, and performance evaluation. It supports both classical and AI-based methods, enabling repeatable, comparable assessments across diverse tasks and configurations. The framework facilitates diagnosis of algorithm behavior, identification of implementation flaws, and selection or tuning of strategies for specific applications. It includes a SysML-based domain-specific language for structured scenario modeling and integrates with the ROS-based system for runtime execution. Validated on patrol, fall assistance, and pick-and-place tasks, the open-source framework is suited for researchers and integrators developing and testing scheduling algorithms under real-world-inspired conditions.
\end{abstract}

\begin{keyword}
robot scheduling algorithms \sep benchmarking tools \sep dynamic environments \sep reinforcement learning
\end{keyword}

\end{frontmatter}

\section*{Metadata}

\begin{table}[!ht]
\footnotesize
\begin{tabular}{|l|p{5.8cm}|p{5.8cm}|}
\hline
\textbf{Nr.} & \textbf{Code metadata description} & \textbf{Please fill in this column} \\
\hline
C1 & Current code version & 1.0.0 \\
\hline
C2 & Permanent link to code/repository used for this code version & \url{https://github.com/RCPRG-ros-pkg/Smit-Sim/tree/v1.0.0} \\
\hline
C3  & Permanent link to Reproducible Capsule & \url{https://github.com/RCPRG-ros-pkg/Smit-Sim/blob/v1.0.0/Dockerfile}\\
\hline
C4 & Legal Code License   & MIT License\\
\hline
C5 & Code versioning system used & git \\
\hline
C6 & Software code languages, tools, and services used & python3.7, bash, ROS melodic\\
\hline
C7 & Compilation requirements, operating environments \& dependencies & \url{https://github.com/RCPRG-ros-pkg/Smit-Sim/blob/main/requirements.txt}\\
\hline
C8 & If available Link to developer documentation/manual & \url{https://github.com/RCPRG-ros-pkg/Smit-Sim/blob/main/README.md} \\
\hline
C9 & Support email for questions & daniel.gieldowski@pw.edu.pl \\
\hline
\end{tabular}
\caption{Code metadata (mandatory)}
\label{codeMetadata} 
\end{table}

\section{Motivation and significance}
\subsection{Problem statement}

The growing demand for robotisation in sectors facing labour shortages highlights the need for autonomous robots capable of managing multiple tasks efficiently. A critical challenge in this context is designing and evaluating scheduling algorithms that can operate reliably in dynamic~\cite{34775}, uncertain~\cite{9942275}, and human-inhabited environments~\cite{FERREIRA2021108094,RicoZmnWut2025}. While AI-based approaches~\cite{shyalika2020reinforcement,TEJER2024107300}, such as reinforcement learning, offer promising solutions, their real-world deployment remains difficult due to the unpredictability of operational contexts and the lack of reliable, comparable evaluation tools.

Existing simulation platforms and benchmarking frameworks, although useful, often fall short in capturing the complexity of real-world conditions and ensuring transparent comparison of algorithmic performance. As a result, the assessment of scheduling algorithms lacks consistency and reproducibility, limiting their integration into practical robotic systems.

The goal of this work is to provide a software framework that enables systematic and reproducible evaluation of both existing and emerging scheduling approaches, offering a solid foundation for the development and comparison of new algorithms. Demonstrating strong performance in some scenarios or failure in others are both valuable outcomes, as they highlight the framework’s ability to reveal algorithmic strengths and weaknesses in dynamic and uncertain environments. To facilitate adoption, we provide a Dockerfile for easy deployment and an introductory video\footnote{\url{https://vimeo.com/1122196556}}
.

\subsection{Software capabilities}
The software framework is dedicated to training and benchmarking robot scheduling algorithms. It enables systematic, repeatable testing in configurable scenarios and ensures fair performance comparisons across algorithms. The framework supports uncertainty modeling, task-specific scenario generation, and comparative evaluation, making it suitable for both algorithm developers and mobile manipulator integrators. It also includes a Domain-Specific Language (DSL) based on Systems Modeling Language (SysML), which facilitates structured documentation and analysis of benchmarking setups using a model-based systems engineering approach. Additionally, the SysML-based description supports assessing software consistency between simulation and physical deployment via SPSysML~\cite{spsysml}.

\begin{figure}[htb]
    \centering
\includegraphics[scale=.65]{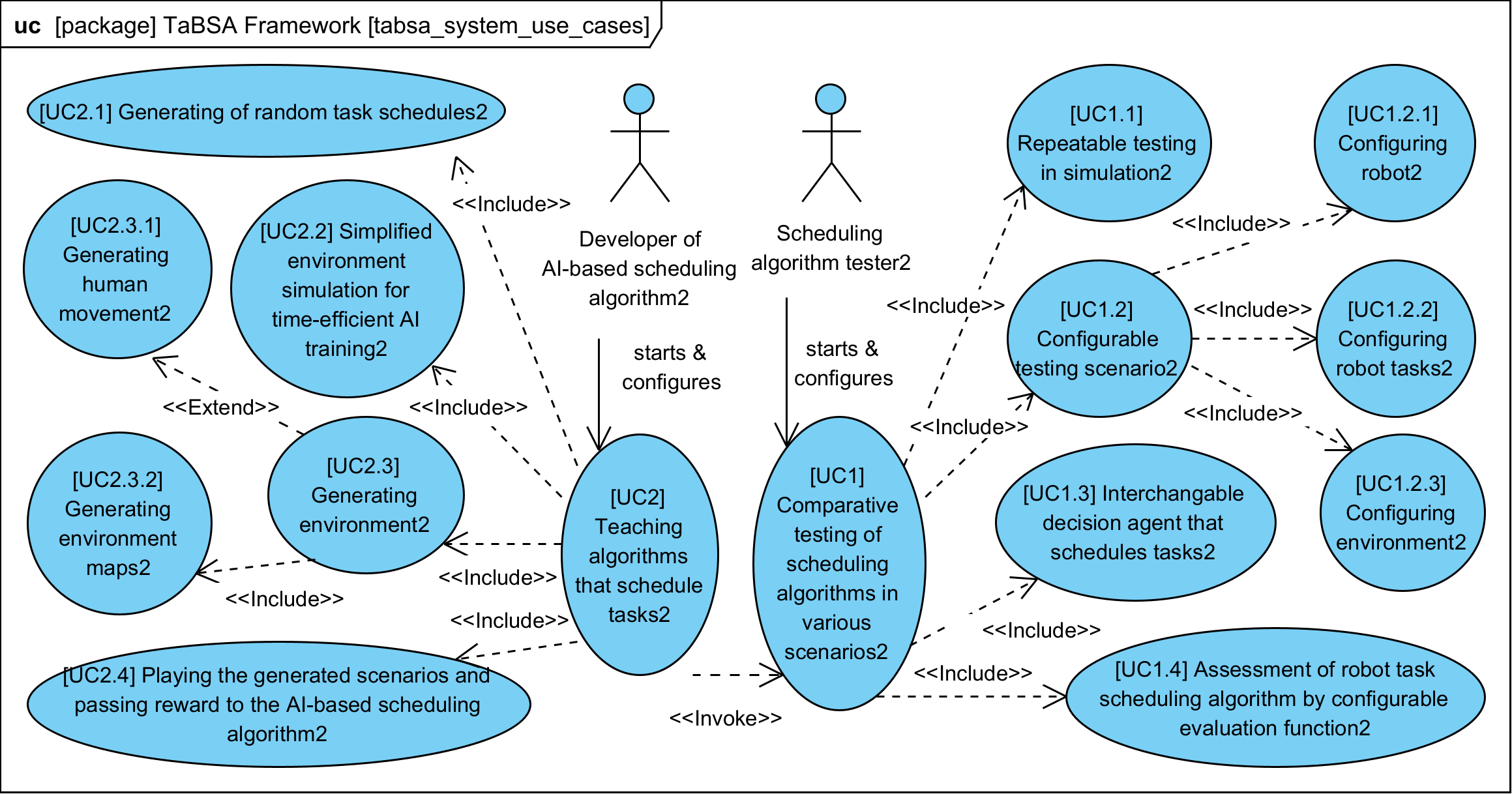}
    \caption{Use cases of the proposed benchmarking system}
    \label{fig:use-case}
\end{figure}

This article presents the framework’s architecture, usage, and validation. It also outlines its integration with robot control systems, including the ROS-based TaskER platform~\cite{9180351} for safe task switching. The proposed framework aims to standardize and accelerate the development of robust, multitasking robots for real-world applications. The framework’s capabilities are illustrated through the use case diagram~\Fig{fig:use-case}.

The proposed framework facilitates benchmarking and development of task scheduling algorithms in realistic robotic environments. It has been validated in three service robot scenarios:
\begin{itemize}
\item patrol and monitoring tasks in dynamic spaces,
\item assistance in human fall detection and intervention,
\item pick-and-place operations involving object manipulation.
\end{itemize}
We provide ready-to-use scenario generators, benchmarking tools, and an interface for training AI-based algorithms under controlled uncertainty. 
 Researchers and integrators can test algorithms in reproducible settings and evaluate their robustness to changing conditions.

The main contribution of this work can be summarized as follows:
\begin{itemize}
\item a complete framework for training and benchmarking robot task scheduling algorithms in uncertain, human-inhabited environments,
\item integration of a domain-specific language for formal documentation and analysis of benchmarking systems,
\item open-source implementation validated in real-world-inspired service robot scenarios, enabling repeatable and comparable evaluation of scheduling methods.
\end{itemize}

\section{Software description}
\begin{figure}[ht]
\begin{subfigure}[b]{0.5\linewidth}
 	\includegraphics[scale=.56]{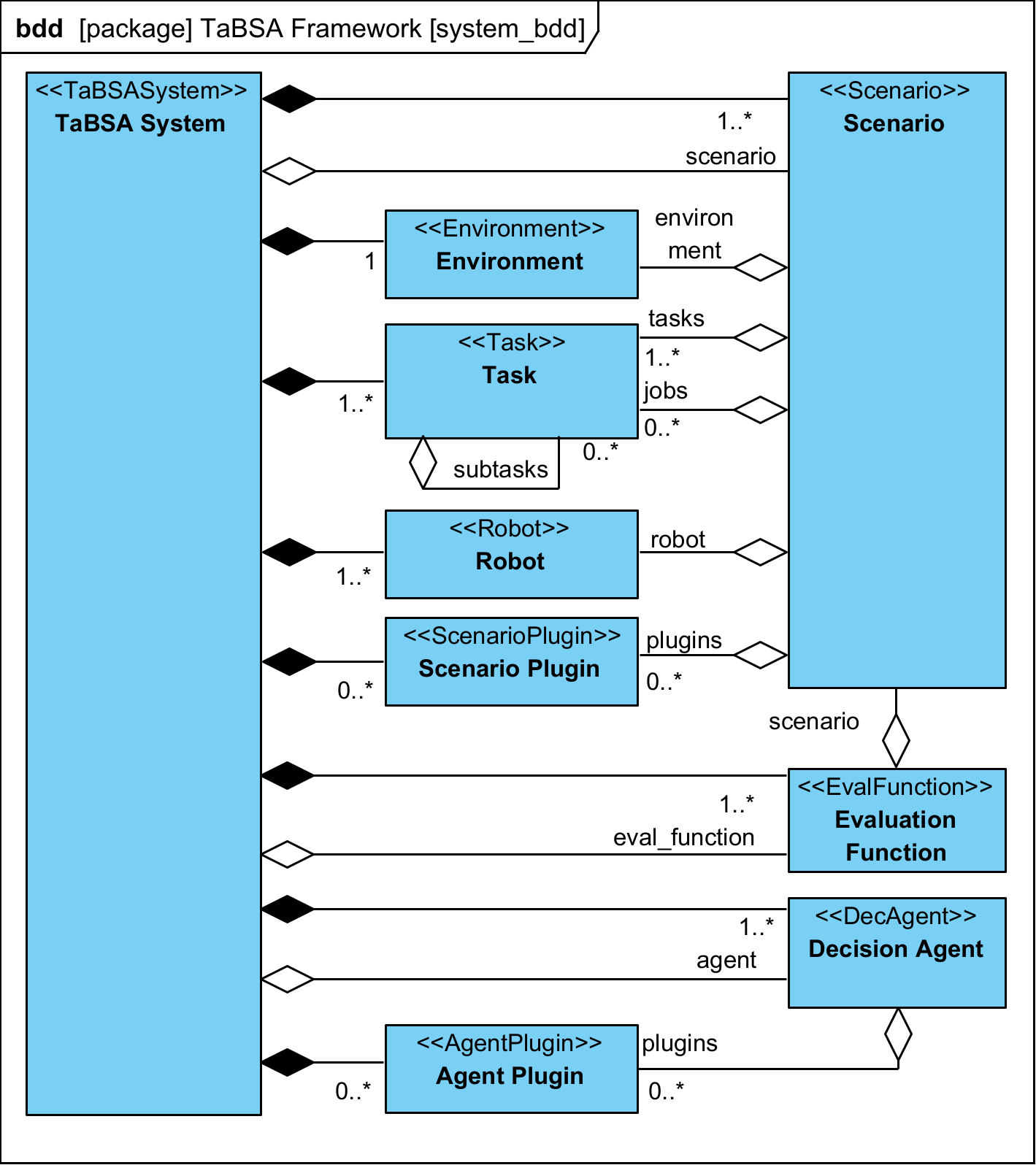}
	\caption{TaBSA system structure}
	\label{fig:smit_system_stereotyped_bdd}
   
\end{subfigure}
\begin{subfigure}[b]{0.5\linewidth}
	\includegraphics[scale=.56]{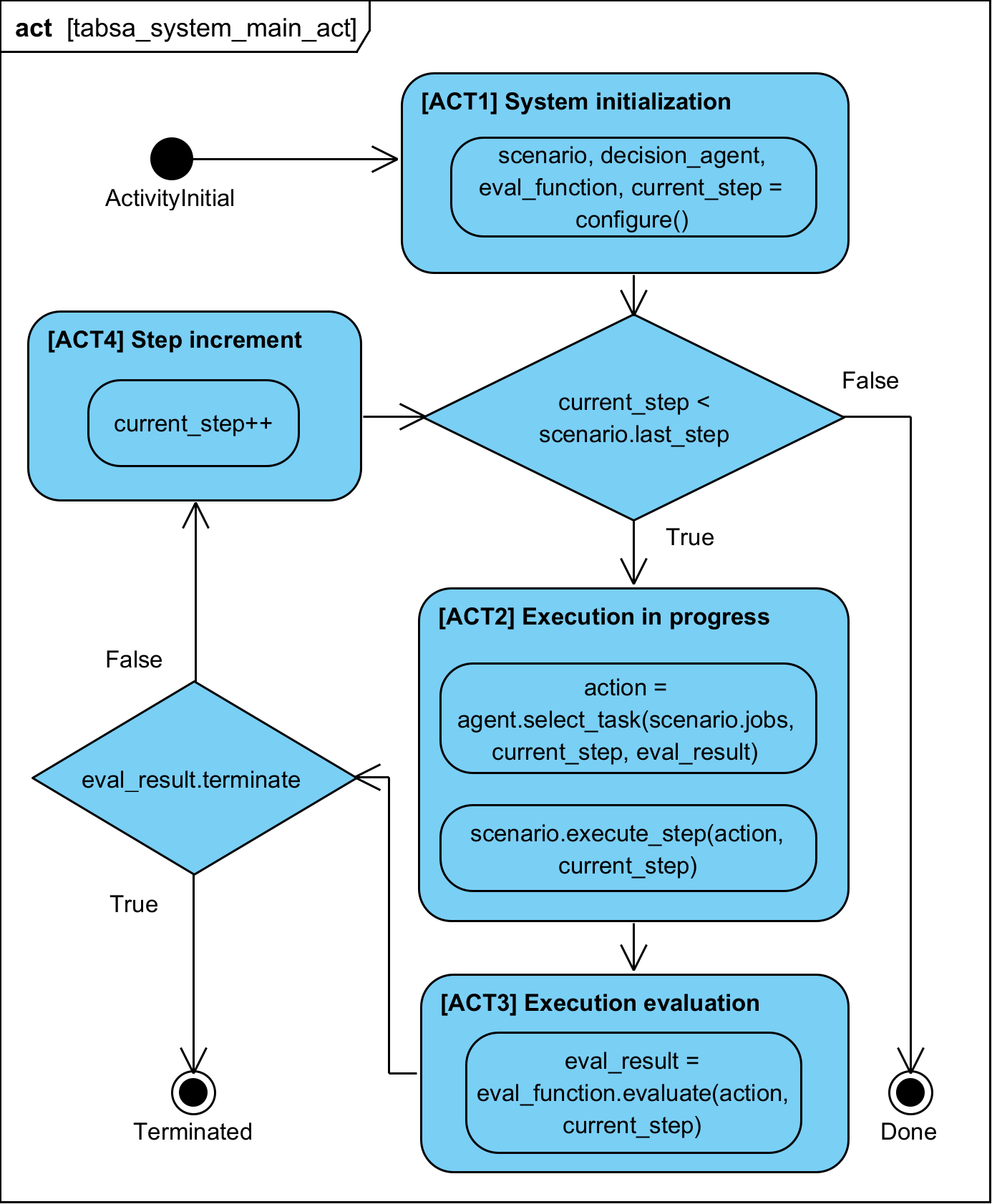}
	\caption{TaBSA system main operation}
	\label{fig:smit_system_main_stereotyped_act}
\end{subfigure}
	\caption{TaBSA structure and behaviour}
	\label{fig:tabsa_struct_beh}
\end{figure}

TaBSA--- the framework for Training and Benchmarking Scheduling Algorithms bases on a metamodel that, by application, gives a specific benchmarking system named TaBSA System. The TaBSA System is organised in a way presented in the SysML block definition diagram in \Fig{fig:smit_system_stereotyped_bdd}.

All system elements that can constitute the variety of configurations are composed (filled rhombus) into \StTabsaSystem{}. The set of the elements for the current configuration (it addresses configurability) of the system is represented by the aggregations (empty rhombus) into the corresponding blocks.

The \StTabsaSystem{} general behaviour is also a part of the metamodel. It is depicted in activity diagram~\ref{fig:smit_system_main_stereotyped_act}. It provides a scheme for repeatable execution of configurable scenarios [UC1]. Activity [ACT1] addresses configurability, [ACT2] in particular addresses teaching and a~step of the task execution [UC2], [ACT3] evaluates the latest task change decision [UC1.4]. A detailed description of the particular operations mentioned in the diagram is presented in the following part of the article.

\subsection{Framework configurability}
\subsubsection{Scope of benchmark scenarios}
\label{scenario-scope}

\begin{figure}[ht]
\centering

	\includegraphics[width=0.6\linewidth]{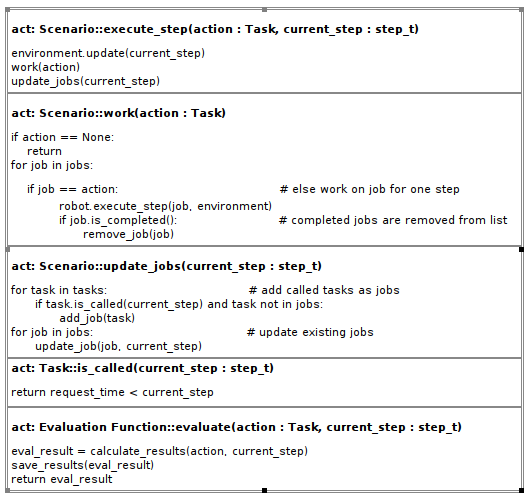}
	\caption{Pseudocodes for some operations of Scenario, Task, and EvalFunction}
	\label{fig:pseudocodes_scenario_task}
\end{figure}

Apart from using a specific \StRobot{}'s model, the \StScenario{} defines important aspects necessary for testing and training task management agents. These are the 'tasks' list and the 'environment'. The \StScenario{} also defines when the simulation starts and ends. Its capabilities can be extended using prepared plugins. The scenario's most important operations are presented in the form of pseudocode in \Fig{fig:pseudocodes_scenario_task}.

The 'tasks' list contains every \StTask{} the robot should work on during the specific training session. We can define the different types of \StTasks{} our robot is supposed to perform as long as they can be represented in our restricted environment, as per [UC1.2.2]. 

The \StScenario{} defines the \StEnvironment{} consisting of a flat map with walls, door openings, and furniture (simple shapes blocking the movement), items placed on the furniture, and moving humans represented by footprints of their legs, as per use case [UC1.2.3]. Simulated humans move on the map using simple kinematic equations and PRM algorithm for path planning, and they execute complex behaviours like going though list of destinations. Humans are embedded on the map, so the robot has to avoid them. Robot manipulation targets placed in the environment are moveable by the robot. These provide a highly simplified representation of a real robot's environment but are sufficient to evaluate \StTask{} execution order and allow training of neural network-based algorithm in a reasonable time. Due to its simplicity, it also satisfies the use case [UC2.2].

\subsection{Tasks adaptation}
Among the operations of \StTask{}, the most important is \textit{work\_for()}, whose implementation determines how the \StTask{} is executed. Some \StTasks{} may be critical, and exceeding their deadline may cause serious consequences. Thus, we define \textit{deathtime} attribute of \StTask{}. \textit{Deathtime} equals:
\begin{equation}
  deathtime = deadline + max\_delay.
\end{equation}
The properties of the example \StTasks{} are as described:
    \begin{multicols}{2}
\begin{itemize}
    \item \textit{effect} -- contains \StTask{}'s current position and state of \StRobot{} and \StEnvironment{} that the \StTask{} sets during execution or if finished. The examples are the \StRobot{}'s localisation or end-position of an object being transported,
    \item \textit{deadline} -- the time during the \StScenario{} by which we would like the \StTask{} to be completed,
    \item \textit{priority} -- a priority of the \StTask{} (required by some \StDecAgents{}), the higher, the more important the task is,
    \item \textit{preemptive} -- defines if the \StTask{} can be interrupted after the \StRobot{} starts to work on it,
    \item \textit{type} -- the name of the \StTask{} type in text form, useful for presenting the system's state and scheduling algorithms assessments,
    \item \textit{estimated\_duration} -- estimated time required to complete the \StTask{} if its execution is started at the current time,
    \item \textit{distance\_from\_robot} -- distance from the \StRobot{}'s position to the \StTask{}'s position,
    \item \textit{deathtime} -- the time at which the task exceeds the deadline by maximum delay. Exceeding this time violates safety; thus, if \textit{deathtime} is up, the \StTask{} is terminated, and the \StScenario{} is failed.
\end{itemize}
\end{multicols}
These properties of \StTask{} are set and updated by the \StScenario{} or its \StScenarioPlugins{} and used by the \StDecAgents{} (to decide which \StTask{} should be performed) and \StEvalFunctions{} (to assess the current situation).

\subsection{Benchmark configurability}
\label{benchmark-configurability}

Apart from \StScenario{} and its \StTasks{} the \StTabsaSystem{} requires other blocks to function, such as \StDecAgent{} and its \StAgentPlugins{} [UC1.3], \StEvalFunction{} [UC1.4], \StScenarioPlugins{}, and \StRobot{} [UC1.2.1]. These blocs depend on specific applications; thus, the users must configure them accordingly. They must implement the following operations and communication interfaces. 

\subsubsection{Configuring the Decision Agents}
The \StDecAgent{} has one basic operation \textit{select\_task()} for selecting a \StTask{} for execution. To make this decision, it receives the list of jobs, the current time, and the output of the \StEvalFunction{} from the previous decision. The job list contains tasks already submitted but not completed (the 'jobs' list of the \StScenario{}). The list may contain more \StTasks{} than the \StDecAgent{} can process. If this is the case, it is at the \StDecAgent{}'s discretion to limit the number of \StTasks{} to be considered. The functionality of the agents can be extended using \StAgentPlugins{}. In one benchmark, you can execute different Decision Agents at once for comparison.

\subsubsection{Configuring the Evaluation Function}

The \StEvalFunction{} implements the \textit{calculate\_results()} operation. This function receives from the system the current time and the \StTask{} selected by the \StDecAgent{} for execution. The type of value returned by the operation is intentionally undefined so that it can be freely extended according to the user's needs. For example, it could contain statistical data on the robot's workflow or a quantitative assessment. The only requirement is to have a 'terminate' boolean variable.

\subsubsection{Creating the Agent Plugins}

An \StAgentPlugin{} extends an agent with a selected capability of considerable complexity. Each \StAgentPlugin{} has its unique interface, depending on how it works. This is due to the infinite number of potential skills with which a \StDecAgent{} can be extended. By separating a capability as an \StAgentPlugin{}, there is no need to implement it individually for each \StDecAgent{}, and new \StDecAgents{} are not forced to expand the code of the old ones to have it. Any \StDecAgent{} can use these \StAgentPlugins{}. An illustrative example of an \StAgentPlugin{} could be a 'task request predictor' that estimates the timestamps of future tasks. 

\subsubsection{Creating the Scenario Plugins}

\StScenarioPlugins{}' role is job processing, facilitating the modification of their parameters according to the \StScenario{} and the \StEnvironment{} as user needs. For this reason, they require implementing a single \textit{update\_job()} operation (run within \StScenario{}'s \textit{update\_job()} operation~--~\Fig{fig:pseudocodes_scenario_task}) that modifies the submitted \StTask{} based on the current time within the \StScenario{}. By design, changes made by \StScenarioPlugin{} should be characterised by significant complexity but potentially not desirable on every run. Separating them allows them to be selectively activated for the current \StScenario{}. \StScenarioPlugins{} may build up extra domain knowledge into the \StTask{} structure. An illustrative example could be a 'task duration estimator' that calculates \StTask{}'s duration estimate so \StDecAgents{} may use it.

\subsubsection{Configuring the Robot}
As the name suggests, the \StRobot{} represents the machine whose work is evaluated during the \StScenario{}. Different \StRobot{} controllers may have different structures and functions. For this reason, we deliberately do not detail the \StRobot{}'s operations and properties. We only require it to have an execute\_step() operation that takes the currently selected \StTask{} to be executed and the \StEnvironment{} as arguments.

\subsection{Test-case generators}
\label{test-case-configuration}
The test case configuration that is the occupancy map, human trajectories, and the real--- unknown to the robot, task schedule, can be set manually for a given specific application, or can be generated with the prepared generators.
\subsubsection{The robot's environment generation}

\begin{figure}[ht]
\begin{subfigure}[b]{0.5\linewidth}

	\includegraphics[scale=.55]{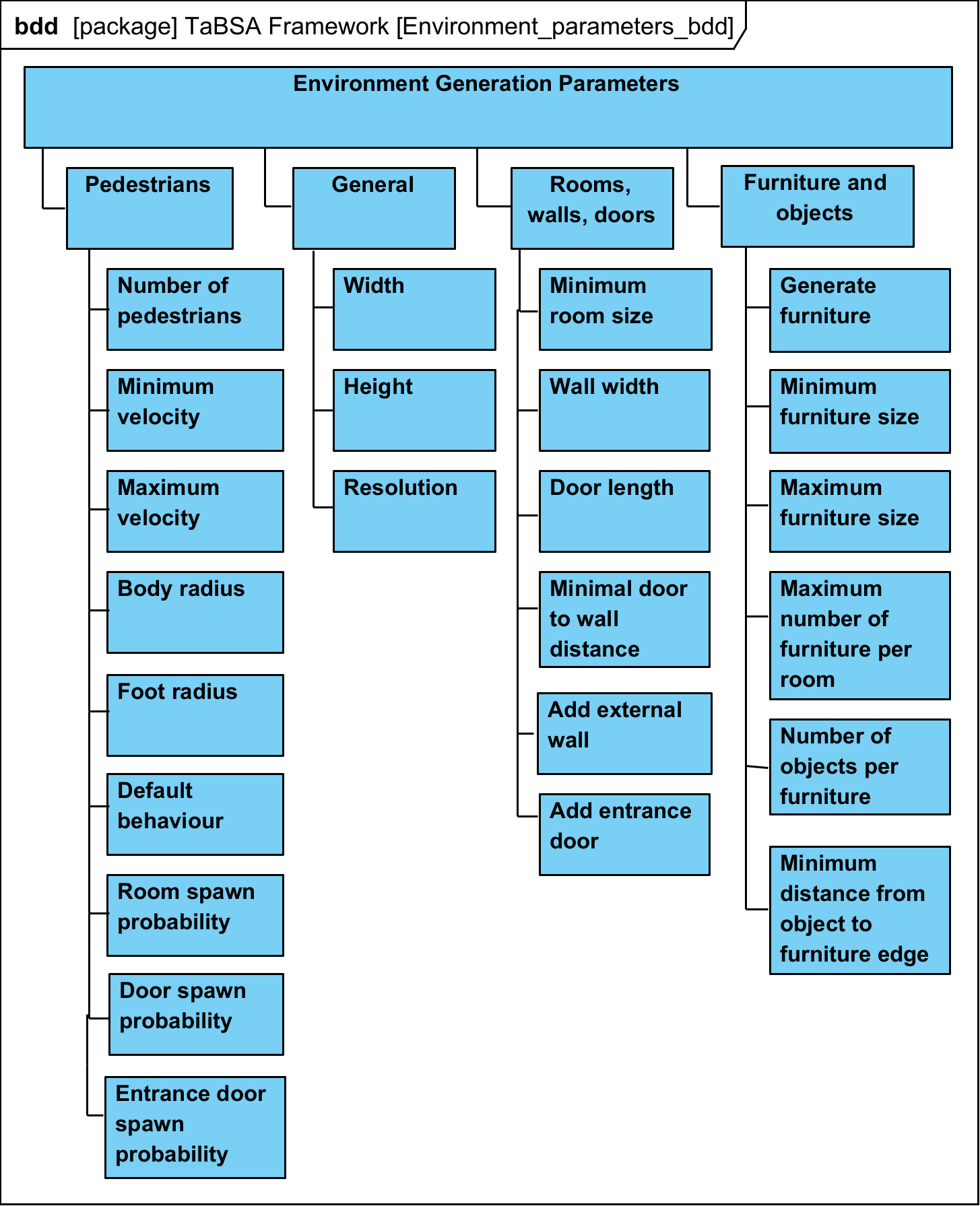}
	\caption{Configurable environment parameters}
	\label{fig:smit_system_env_params}
\end{subfigure}
\begin{subfigure}[b]{0.5\linewidth}
    \centering
    \includegraphics[width=\linewidth]{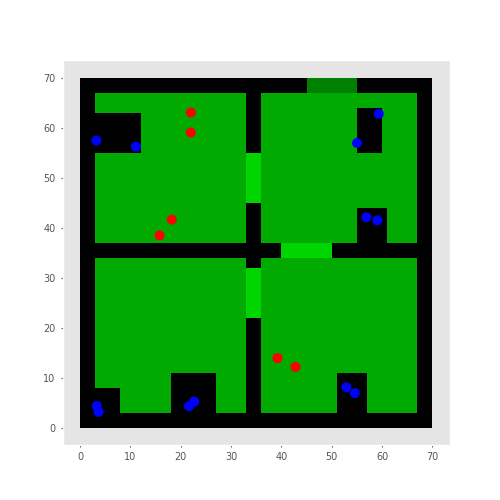}
    \caption{Example randomly generated map with objects (blue), calculated probability distribution for pedestrian spawning (the darker green the bigger probability), and example footprints of the pedestrians (red) at the system runtime}
    \label{fig:sample_random_map_with_objects}
\end{subfigure}

	\caption{Robot's environment configuration}
	\label{fig:smit_env_configuration}
\end{figure}

The \StEnvironment{} in which the \StRobot{} works is \StScenario{}'s property. The static obstacles emulating walls for the test session are generated using the recursive division method configured with the provided parameters (\Fig{fig:smit_system_env_params}). First, all rooms, walls, and doors are generated considering the provided constraints. After that, the number of pieces of furniture is added to the map so as not to stand in front of any door. The number of furniture pieces is limited but not constant, as some rooms may be too small to place that much furniture, or the furniture just won't generate properly in a finite number of tries. A set number of manipulation-intended items is placed on each piece of furniture. Ultimately, we generate starting and destination poses for human actors and define their behaviour.

The probability map is calculated based on the pedestrian spawning probabilities set during the environment configuration~(\Fig{fig:smit_env_configuration}b\textit{/Pedestrians}). Different probabilities can be set for the rooms, doors, and map entrance. The example environment map is presented in \Fig{fig:sample_random_map_with_objects}. All of this satisfy [UC2.3], specifically [UC2.3.1] and [UC2.3.2].

\subsubsection{Task generation}

The \StTasks{} implemented in the \StScenario{} are randomly generated using a chosen seed. This allows us to recreate the \StTask{} configurations while testing different \StDecAgents{} or plugin configurations without saving all the \StTask{} information, as per [UC2.1]. While simulating the \StRobot{}'s work in a specific \StEnvironment{}, different seeds may represent different timeframes during the day or even certain days of the week, month, or year. For each task, we generate the time when it is given to the robot, the deadline, and the necessary environment positions -- for example, the robot's target position or positions of potential manipulation objects. The times must fit between the first step of the execution and the last step of the scenario. The positions are generated while considering the obstacles in the \StEnvironment{}. For example, the \StRobot{}'s position should be able to navigate to, and the object's position should be within one of the existing pieces of furniture.

\subsection{Example benchmarking sessions}
\label{example-benchmarking}
\subsubsection{Decision agent training}

As mentioned before, DQN \StDecAgent{} requires a trained neural network to operate. Training this network may be considered an example of a successful benchmarking session utilising our Mobile Manipulator System. The \StDecAgent{} was taught using reinforcement learning (DQN algorithm). The training consisted of the \StDecAgent{} rehearsing numerous \StScenarios{}. During the training, the \StScenarios{} reflected 4 hours of social \StRobot{} work, during which the \StDecAgent{} was expected to complete 12 \StTasks{} of each of the 3 types. In each, the \StTasks{} were generated using a new seed. The DQN \StDecAgent{}'s actions were evaluated using DQN Eval, and the reward received during this stage was used as a reward for reinforcement learning. Learning took several million steps, each representing 5 seconds of virtual time. Individual \StScenarios{} ended if all \StTasks{} were completed, one of the \StTasks{} died or \StTask{} completion exceeded the \StScenario{} time. Once training was complete, the network was saved for further use. The training was repeated several times to verify differences in the performance of the different network structures and the value of the reward for individual actions.

\section{Illustrative example}

\begin{figure}[ht]
\begin{subfigure}[b]{0.5\linewidth}

    \centering
    \includegraphics[scale=.62]{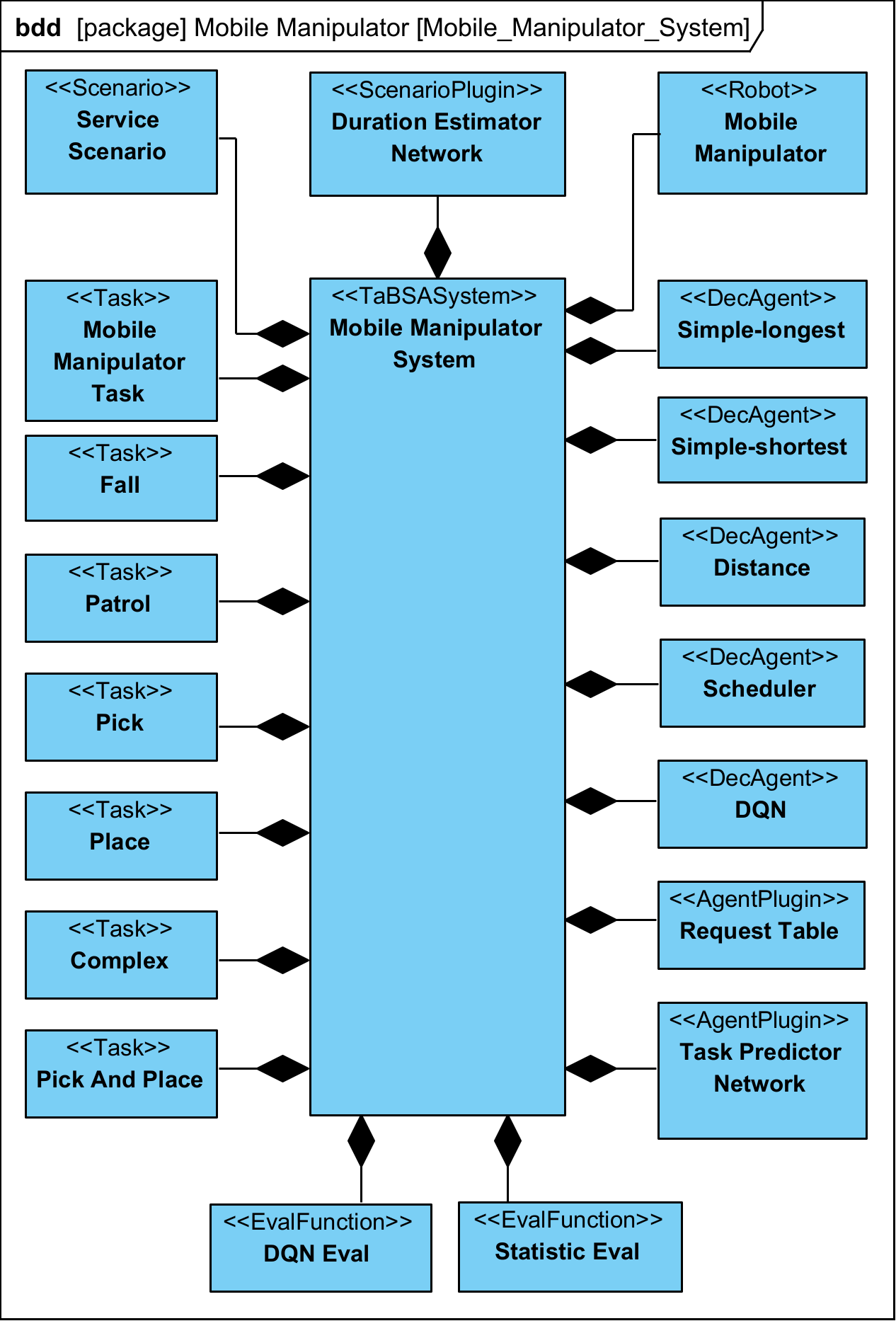}
    \caption{Mobile Manipulator System}
    \label{fig:mobile_manipulator_system}
\end{subfigure}
\begin{subfigure}[b]{0.5\linewidth}
    \centering
    \includegraphics[scale=.62]{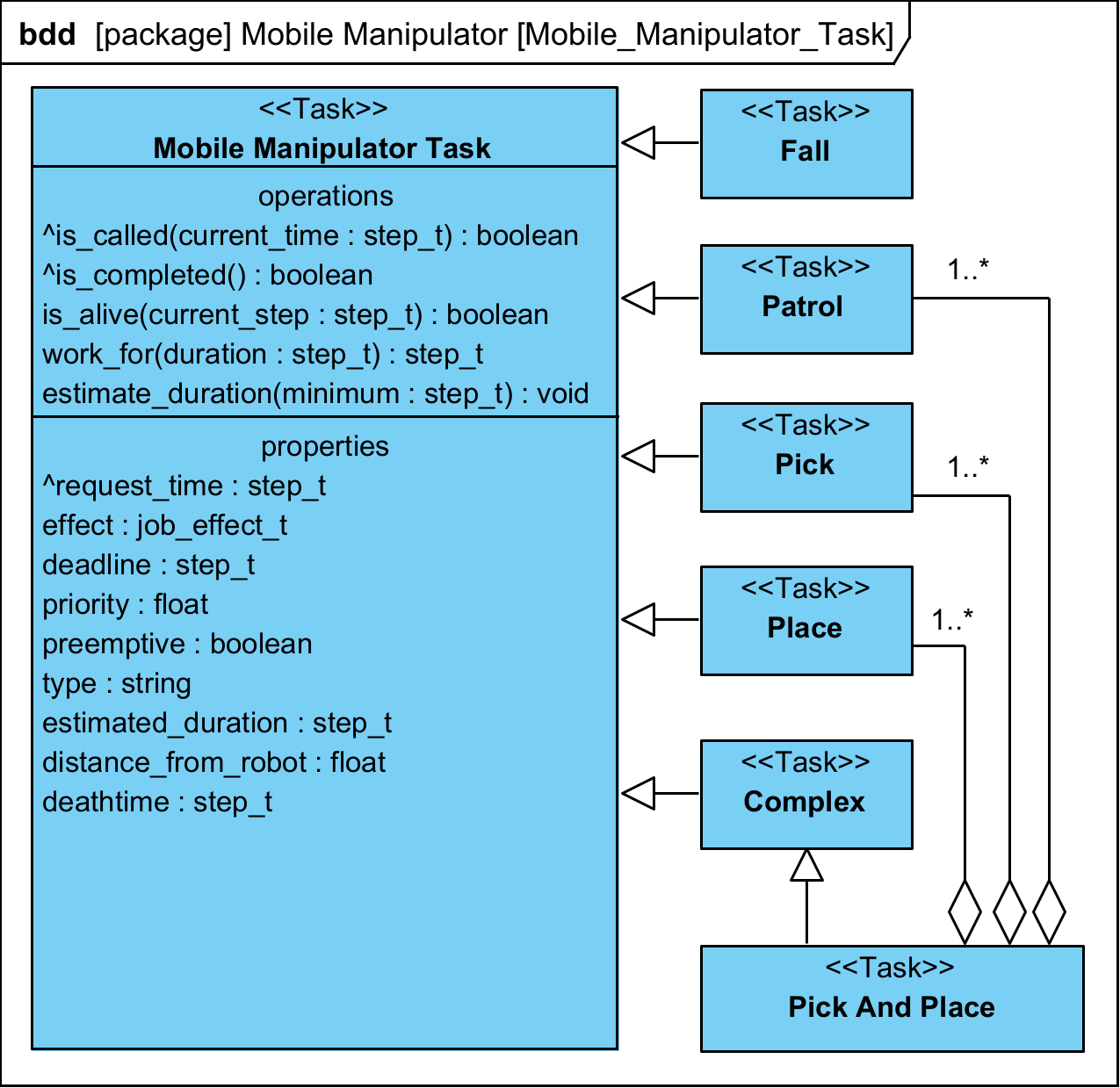}
    \caption{Mobile Manipulator Task}
    \label{fig:mobile_manipulator_task}
\end{subfigure}
	\caption{Mobile Manipulator system definition}
	\label{fig:mobile_manipulator_system_definition}
\end{figure}

In the following section, we describe an example Mobile Manipulator System based on the \StTabsaSystem{}. It contains its own components based on the \StScenario{}, \StScenarioPlugin{}, \StTask{}, \StRobot{}, \StDecAgent{}, \StAgentPlugin{}, and \StEvalFunction{} stereotypes. They are graphically presented in \Fig{fig:mobile_manipulator_system}.

\subsection{Tasks}

Mobile Manipulator Task is based on the \StTask{} stereotype but implements numerous additional operations and properties (see \Fig{fig:mobile_manipulator_task}).

To represent tasks composed of several simpler tasks, we prepared the Complex \StTask{}. This task possesses a list of 'subtasks' that are worked on one after another. For this reason, the \StTask{}'s 'estimated\_duration' equals $\sum\limits_{subtasks}(subtask.estimated\_duration)$. The \StTask{}'s 'priority' was experimentally set to $\max(subtask.priority: subtasks)$ to the sum of priorities of subtasks. The only composite \StTask{} prepared within the system was Pick And Place. This \StTask{} aims to transport an item from one place to another. 

\subsection{Decision Agents}
\begin{itemize}

\item \texttt{Simple-longest} and \texttt{Simple-shortest} \StDecAgents{} select the longest and the shortest \StTask{} for the execution, respectively. 
\item The Distance \StDecAgent{} chooses the \StTask{} that is the furthest from the \StRobot{}. 
\item The Scheduler \StDecAgent{} uses the Request Table \StAgentPlugin{}. Request Table sorts the received \StTasks{} by highest 'priority'. \StTasks{} with the same 'priority' are then sorted by the earliest 'request\_time'.
\item DQN \StDecAgent{} implements the neural network trained using the Deep Q Network~\cite{NIPS2016_8d8818c8} (DQN) algorithm to choose the \StTask{} for execution.
\end{itemize}

\subsection{Scenario}
The example Service Scenario, based on the \StScenario{} stereotype, has a Mobile Manipulator \StRobot{}, a single \StScenarioPlugin{}~-- Duration Estimator Network, and only Mobile Manipulator Task \StTasks{}. The \StScenario{} is meant to represent several hours of work for the mobile service \StRobot{}. The \StRobot{} has a set velocity for traveling between \StTasks{}. A certain number of \StTasks{} of each type -- Fall, Patrol, and Pick And Place~-- is generated within these hours. The properties of these \StTasks{} are updated every time by the update\_job() operation: \textit{estimate\_duration()} is called, 'distance\_from\_robot' is calculated, and every path that needs to be planned within the \StTask{} is refreshed using the current state of the environment.

\subsection{Evaluation functions}

Two \StEvalFunctions{} are currently implemented within the Mobile Manipulator System: DQN Eval and Statistic Eval. Both of them have their specific purposes.

DQN Eval is the \StEvalFunction{} created mainly to train the DQN \StDecAgent{} because the \StDecAgent{}'s most important part is the neural network trained using reinforcement learning. It requires a proper reward function to learn how it should operate. Therefore, DQN Eval functions as both an \StEvalFunction{} for the \StTabsaSystem{} and a training reward for DQN \StDecAgent{}. DQN Eval returns a 'terminate' flag if all \StTasks{} are completed or one of the \StTasks{} is terminated. It also returns the 'reward' parameter that depends on the \StDecAgent{}'s performance. If the \StDecAgent{} chooses an existing \StTask{}, one or all \StTasks{} are completed, the reward is positive. If the \StTask{} chosen by the \StDecAgent{} doesn't exist or one of the \StTasks{} is terminated, the reward takes the form of a negative penalty. The \StDecAgent{} is also slightly penalized for switching \StTasks{} they work on to discourage it from jumping between \StTasks{} too often. All of these values are represented by appropriately named parameters. The reward function is presented in the equation \eqref{eq:dqn_eval_reward}. The additional parameter--- 'penalty\_change\_job' is added to the reward only in the last two cases if the current action is different than the 'previous\_action' property.

\begin{equation}\footnotesize
    reward = \left\{ \begin{array}{ll}
         reward\_all\_complete & if\,all\,tasks\,are\\&\quad completed \\
         penalty\_dead\_job & if\,at\,least\,one\,job\\&\quad is\,dead \\
         reward\_job\_complete & if\,job\,was\,just\\&\quad completed \\
         0 & if\,there\,is\,no\,jobs \\&\quad in\,the\,list\\
         reward\_real\_job & if\,real\,job\,was\\&\quad selected \\
         penalty\_nonexistent\_job & if\,nonexistent\,job\\&\quad was\,chosen \\
    \end{array} \right.
    \label{eq:dqn_eval_reward}
\end{equation}

Statistic Eval aims to numerically assess the work of any \StDecAgent{} it monitors using statistics. This makes it good for every \StDecAgent{} in the \StTabsaSystem{}. It allows the user to verify what the decisions made by the \StDecAgent{} accomplish in the chosen \StScenario{}. Statistic Eval returns the 'terminate' flag if all \StTasks{} are completed, one of the \StTasks{} is terminated or the \StDecAgent{} switches between \StTasks{} and returns to the same one for the third time within 3 minutes. In the documentation, there are other useful values returned by the evaluation function on each iteration.

\subsection{Experimental results}

Initially, 50 \StScenarios{} with random \StTasks{} were generated for each \StDecAgent{} under test. The \StScenarios{} lasted 4 hours and contained 12 \StTasks{} of each of the three types. This time the Statistic Eval was used to evaluate the performance of the \StDecAgents{}. The results of each run were saved for analysis. Some of the results for six different \StDecAgents{} (Distance with 'ratio' 0.5, two versions of DQN, Scheduler, \texttt{Simple-longest}with 'hesitance' 0.5, and \texttt{Simple-shortest} with 'hesitance' 0.5) are shown below, as they presented us with useful information about the \StDecAgents{} and allowed us to improve them.

\begin{figure}[ht]
    \centering
    \includegraphics[width=0.5\linewidth]{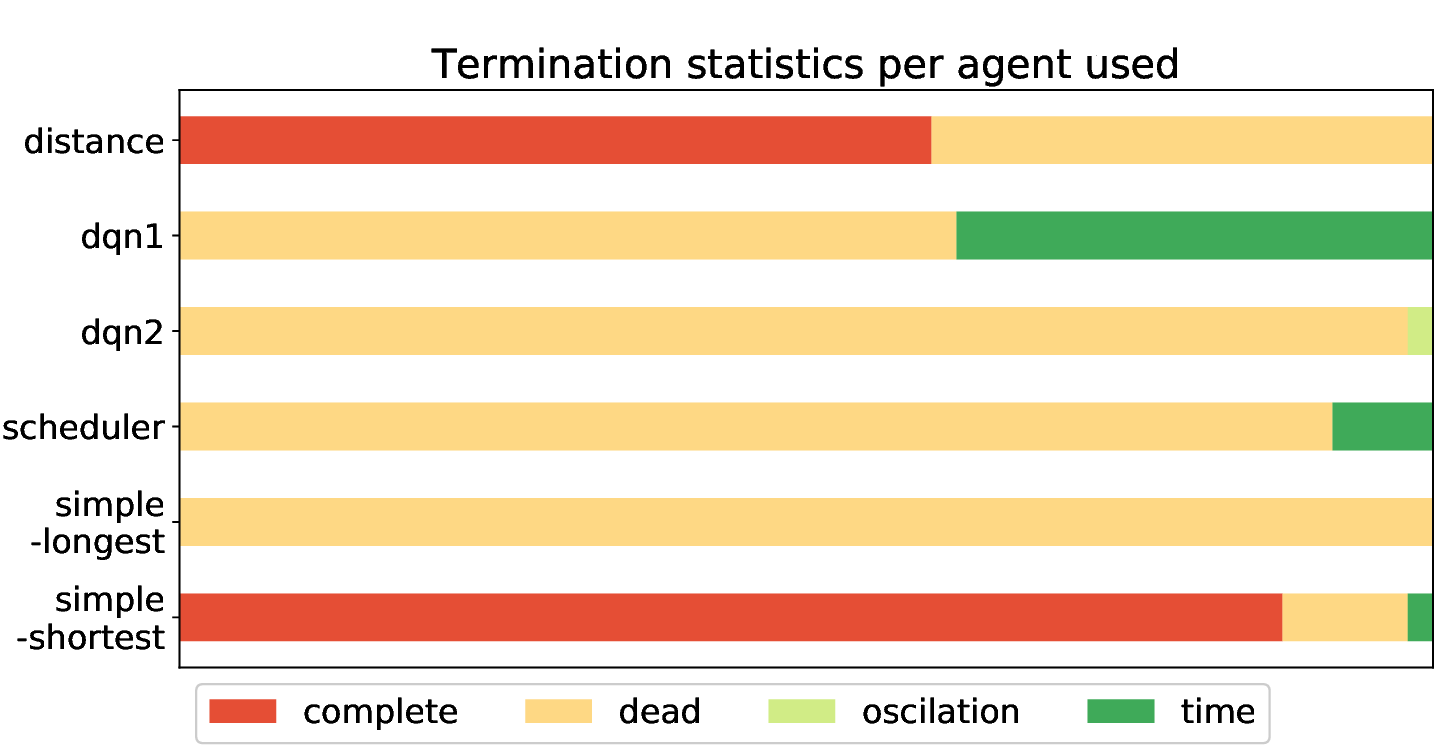}
    \caption{Termination statistics for 6 \StDecAgents{} (Distance, two versions of DQN, Scheduler, Simple, and \texttt{Simple-shortest})}
    \label{fig:termination_results}
\end{figure}

\Fig{fig:termination_results} includes statistics for scenario termination cause for different types of \StDecAgents{} presented as bars. We can see that \texttt{Simple-shortest} and Distance \StDecAgents{} were moderately successful in completing more than half of the \StScenarios{}. Scheduler and \texttt{Simple-longest} \StDecAgents{} both failed for different reasons. During the framework execution, we identified unwanted behaviour of the first DQN \StDecAgent{}. It has terminated a lot of \StScenarios{} due to running out of time. This fact prompted us to reevaluate the code and discover that the penalty for doing nothing was indeed wrongly applied as a positive number. This was fixed for the second network, which, as one can see, doesn't display the same behaviour.

\begin{figure}[ht]
\begin{subfigure}[b]{0.48\linewidth}

    \centering
    \includegraphics[width=\linewidth]{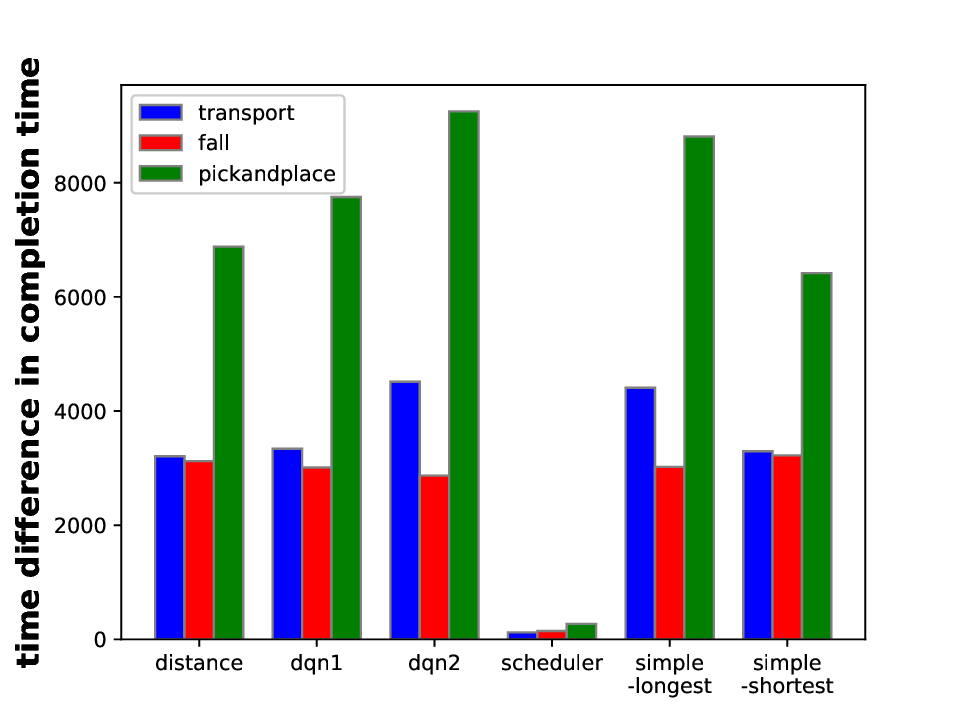}
    \caption{Difference between real completion time and the 'deadline' for different \StTask{} types for different \StDecAgents{}}
    \label{fig:completion_time_difference}
\end{subfigure}\hspace{.2cm}
\begin{subfigure}[b]{0.48\linewidth}
    \centering
    \includegraphics[width=\linewidth]{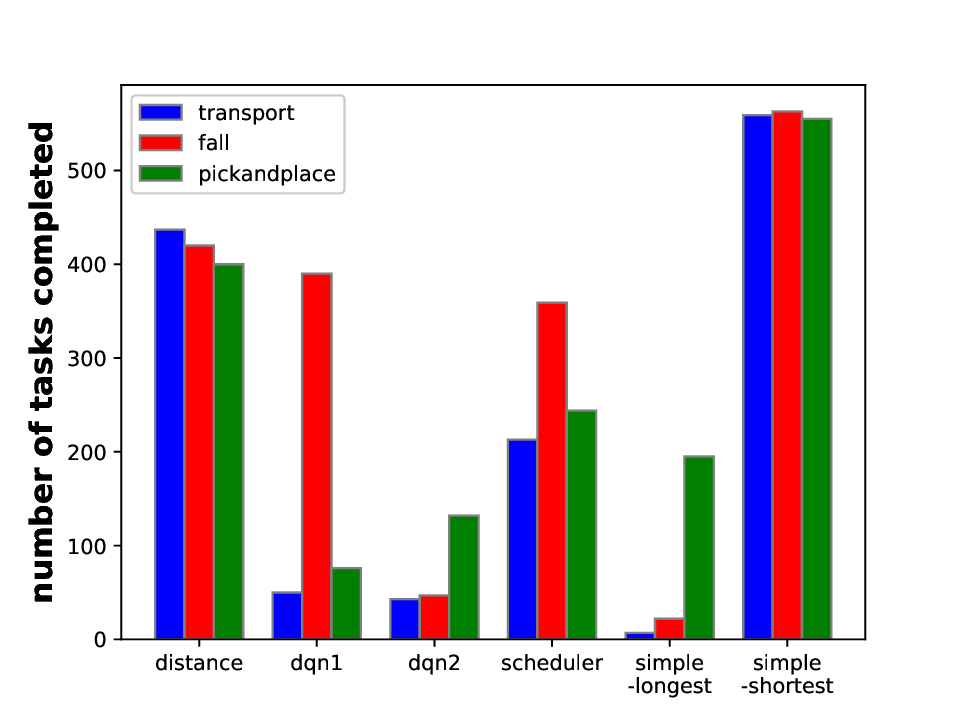}
    \caption{Number of completed \StTasks{} per type for different \StDecAgents{}\\}
    \label{fig:completed_tasks}
\end{subfigure}
	\caption{Comparison of the scheduling algorithms}
	\label{fig:smit_system_agent_comparison}
\end{figure}

\Fig{fig:completion_time_difference} shows the absolute difference between \StTasks{}' completion times and their original 'deadlines' sorted by \StTask{} type. As can be seen, one of the \StDecAgents{} seems to be much better than the others - Scheduler - as it displayed much smaller disparities. This means that \StTasks{} worked on by this agent were completed close to the original 'deadlines', which may be a desirable behaviour in some cases.

\Fig{fig:completed_tasks} shows the number of \StTasks{} of each type completed by each \StDecAgent{} during the 50 \StScenario{}s. These plots also confirm that most \StDecAgents{} work as intended despite failing the \StDecAgents{}. For example, the Scheduler has more Fall \StTasks{} completed than other types, as they have the highest manual 'priority'. On the other hand, \texttt{Simple-longest} chooses to work on the longest \StTasks{} first, so it prioritizes Pick And Place \StTasks{}. The biggest surprise is the first DQN, which seems to have completed almost as many fall \StTasks{} as the Distance \StDecAgent{} but didn't complete any \StScenarios{}. While the DQN-based decision agent performed poorly in our experiments, this outcome highlights a central feature of the proposed framework: its ability to reveal limitations of specific approaches under controlled, reproducible conditions. The framework is designed precisely for such evaluations, enabling algorithm developers to identify, analyze, and address weaknesses in their methods. Thus, the weaker performance of DQN in this setting demonstrates the effectiveness of the proposed framework in revealing algorithmic limitations or providing valuable insights for further improvements.

\begin{figure}[ht]

\begin{subfigure}[b]{0.48\linewidth}
    \centering
   \includegraphics[width=\linewidth]{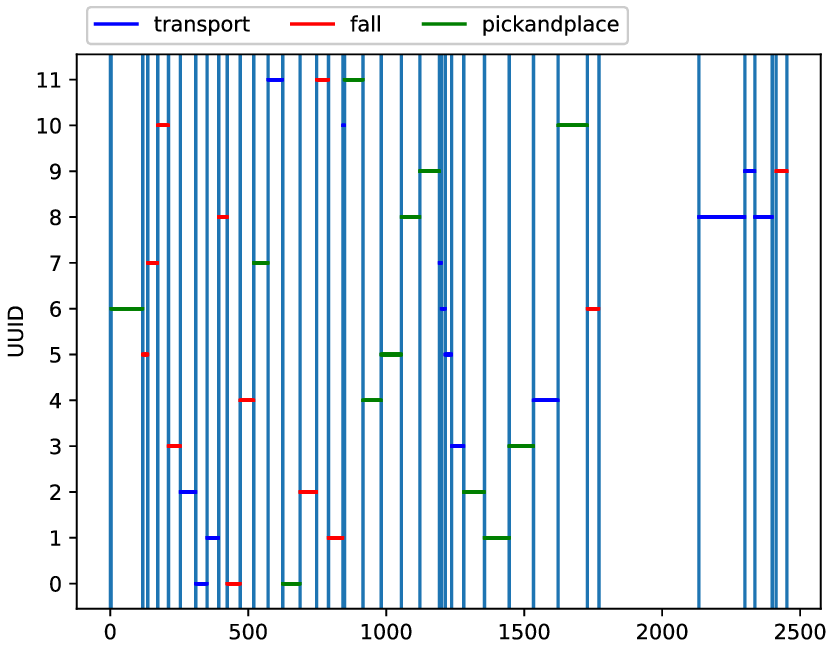}
   \caption{Scenario course as performed by \texttt{Simple-shortest} \StDecAgent{}}
    \label{fig:course_simple2}
\end{subfigure}\hspace{.2cm}
\begin{subfigure}[b]{0.48\linewidth}
    \centering
    \includegraphics[width=\linewidth]{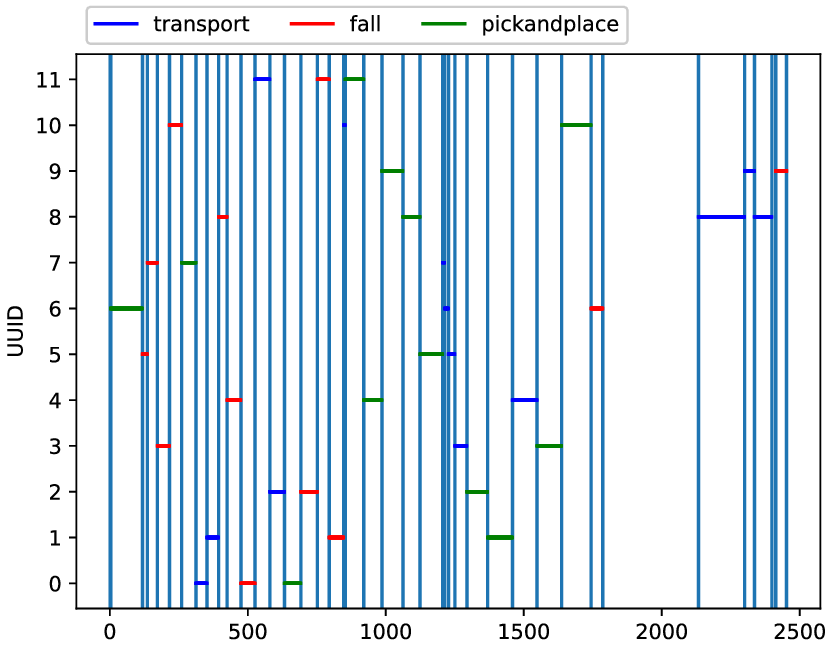}
    \caption{Scenario course as performed by Distance \StDecAgent{}}
    \label{fig:course_distance}
\end{subfigure}
	\caption{Comparison of the example scenario execution}
	\label{fig:smit_system_scenario_execution}
\end{figure}

After that, another experiment was performed, in which each \StDecAgent{} completed the same 100 \StScenarios{} (same seed used for \StTask{} generation). This allowed us to inspect the differences in \StDecAgents{}' behaviours closely. Figures \ref{fig:course_simple2} and \ref{fig:course_distance} show how the same scenario was completed by \texttt{Simple-shortest} and Distance \StDecAgents{}. The horizontal axis represents the passage of time, while the currently performed \StTask{} is identified by its colour and order in the \StTask{} list of the \StScenario{}. For example, a green line at 6 represents the sixth generated Pick And Place task. Comparing the plots, we can see that the \StTasks{} were completed in a slightly different order at the beginning due to differences in decision-making. On the other hand, since about 1700th second, the outcomes are almost identical, probably because these \StTasks{} were given to the agents at the same time late into the day.

\begin{figure}[ht]
\end{figure}

\begin{figure}[ht]
\end{figure}

\section{Impact}
The presented framework has demonstrated significant utility in the quantitative evaluation of scheduling algorithms, referred to as Decision Agents, for service robots operating in dynamic environments. By executing each agent within identical, simulated scenarios that include randomized human movement and diverse tasks, the framework enables fair, reproducible comparisons. It supports both broad benchmarking across general conditions and targeted testing in application-specific setups. The experiments confirmed the framework's capability to reveal subtle behavioral differences between agents, highlight performance patterns, and, crucially, expose implementation or configuration errors—such as misapplied penalties in one of the tested DQN agents. Through detailed statistical outputs and visual comparisons, users can assess algorithm versatility or select the most suitable scheduling strategy for a given robotic application, thereby accelerating both algorithm development and deployment-readiness.

\section{Conclusions}
We proposed and validated a practical benchmarking framework for evaluating scheduling algorithms in mobile service robotic applications. The framework enables configurable simulation of environments, tasks, robot kinematics, and human activity, allowing reproducible and comparative testing of decision-making algorithms under identical conditions. The experiments demonstrated the framework’s effectiveness in identifying both behavioral characteristics and implementation flaws, highlighting its diagnostic value. By visualizing differences in task execution, completion timing, and failure causes, the framework supports both algorithm tuning and application-driven selection. The presented results confirm that the framework can effectively benchmark diverse decision-making agents, from heuristic schedulers to learning-based approaches. The fact that certain algorithms, such as DQN, underperformed demonstrates the framework’s usefulness. It provides a fair and transparent way to assess algorithms and to identify cases where they can be further refined. 

It is open-source and extensible\footnote{\url{https://github.com/RCPRG-ros-pkg/Smit-Sim}}, therefore it serves as a ready-to-use tool for practitioners aiming to improve scheduling strategies or adapt them to specific robot deployments. Finally, the insights gained suggest that task-deadline awareness and preference for shorter tasks increase scheduling robustness in realistic scenarios, while future work should target semantic-level planning for more complex robot activities.

\section*{Acknowledgements}
The research was funded by the Centre for Priority Research Area Artificial Intelligence and Robotics of Warsaw University of Technology, Poland, within the Excellence Initiative: Research University (IDUB) programme, agreement no. 1820/336/Z01/POB2/2021. The work of Kamil Młodzikowski and Dominik Belter was supported by the National Science Centre, Poland, under research project no  UMO-2023/51/B/ST6/01646.

\bibliographystyle{elsarticle-num} 
\bibliography{tamatto}

\begin{thebibliography}{1}
\expandafter\ifx\csname url\endcsname\relax
  \def\url#1{\texttt{#1}}\fi
\expandafter\ifx\csname urlprefix\endcsname\relax\def\urlprefix{URL }\fi
\expandafter\ifx\csname href\endcsname\relax
  \def\href#1#2{#2} \def\path#1{#1}\fi

\bibitem{34775}
S.~Dubowsky, T.~Blubaugh, Planning time-optimal robotic manipulator motions and
  work places for point-to-point tasks, IEEE Transactions on Robotics and
  Automation 5~(3) (1989) 377--381.
\newblock \href {http://dx.doi.org/10.1109/70.34775}
  {\path{doi:10.1109/70.34775}}.

\bibitem{9942275}
B.~Fu, W.~Smith, D.~M. Rizzo, M.~Castanier, M.~Ghaffari, K.~Barton, Robust task
  scheduling for heterogeneous robot teams under capability uncertainty, IEEE
  Transactions on Robotics 39~(2) (2023) 1087--1105.
\newblock \href {http://dx.doi.org/10.1109/TRO.2022.3216068}
  {\path{doi:10.1109/TRO.2022.3216068}}.

\bibitem{FERREIRA2021108094}
C.~Ferreira, G.~Figueira, P.~Amorim, Scheduling human-robot teams in
  collaborative working cells, International Journal of Production Economics
  235 (2021) 108094.
\newblock \href {http://dx.doi.org/10.1016/j.ijpe.2021.108094}
  {\path{doi:10.1016/j.ijpe.2021.108094}}.

\bibitem{RicoZmnWut2025}
T.~Winiarski, D.~Giełdowski, K.~Giełdowski, M.~Tulik, M.~Kobylecka,
  J.~Kunikowska, Concepts for the use of assistive robots and artificial
  intelligence in a nuclear medicine facility, Clinical and Translational
  Imaging\href {http://dx.doi.org/10.1007/s40336-025-00718-8}
  {\path{doi:10.1007/s40336-025-00718-8}}.

\bibitem{shyalika2020reinforcement}
C.~Shyalika, T.~Silva, A.~Karunananda, Reinforcement learning in dynamic task
  scheduling: A review, SN Computer Science 1~(6) (2020) 306.

\bibitem{TEJER2024107300}
M.~Tejer, R.~Szczepanski, T.~Tarczewski, Robust and efficient task scheduling
  for robotics applications with reinforcement learning, Engineering
  Applications of Artificial Intelligence 127 (2024) 107300.
\newblock \href {http://dx.doi.org/10.1016/j.engappai.2023.107300}
  {\path{doi:10.1016/j.engappai.2023.107300}}.

\bibitem{spsysml}
W.~Dudek, N.~Miguel, T.~Winiarski, A sysml-based language for evaluating the
  integrity of simulation and physical embodiments of cyber–physical systems,
  Robotics and Autonomous Systems 185 (2025) 104884.
\newblock \href {http://dx.doi.org/https://doi.org/10.1016/j.robot.2024.104884}
  {\path{doi:https://doi.org/10.1016/j.robot.2024.104884}}.

\bibitem{9180351}
W.~Dudek, T.~Winiarski, Scheduling of a robot’s tasks with the tasker
  framework, IEEE Access 8 (2020) 161449--161471.
\newblock \href {http://dx.doi.org/10.1109/ACCESS.2020.3020265}
  {\path{doi:10.1109/ACCESS.2020.3020265}}.

\bibitem{NIPS2016_8d8818c8}
I.~Osband, C.~Blundell, A.~Pritzel, B.~Van~Roy,
  \href{https://proceedings.neurips.cc/paper_files/paper/2016/file/8d8818c8e140c64c743113f563cf750f-Paper.pdf}{Deep
  exploration via bootstrapped dqn}, in: D.~Lee, M.~Sugiyama, U.~Luxburg,
  I.~Guyon, R.~Garnett (Eds.), Advances in Neural Information Processing
  Systems, Vol.~29, Curran Associates, Inc., 2016.
\newline\urlprefix\url{https://proceedings.neurips.cc/paper_files/paper/2016/file/8d8818c8e140c64c743113f563cf750f-Paper.pdf}

\end{thebibliography}

\end{document}